\documentclass[letterpaper]{article}
\usepackage[preprint]{aaai2027}

\usepackage[hyphens]{url}
\usepackage{graphicx}
\usepackage{natbib}
\usepackage{caption}
\usepackage{algorithm}
\usepackage{algorithmic}
\usepackage{booktabs}
\usepackage{colortbl}
\usepackage{arydshln}
\usepackage{amsmath}

\newcommand{\method}{LoSA}

\title{\method{}: Near-Lossless Sparse Attention for Training-Free Video Diffusion Acceleration}
\author{
    Enhuai Liu\textsuperscript{\rm 1},
    Yunke Wang\textsuperscript{\rm 1},
    Yutong Wang\textsuperscript{\rm 1},
    Changming Sun\textsuperscript{\rm 2},
    Chang Xu\textsuperscript{\rm 1}
}
\affiliations{
    \textsuperscript{\rm 1}The University of Sydney\\
    \textsuperscript{\rm 2}CSIRO
}

\begin{document}

\maketitle

\begin{abstract}
Video diffusion transformers are costly to sample: every denoising step applies
self-attention over a long 3D token sequence, a quadratic cost that
dominates as resolution and duration grow. Sparse attention reduces this cost
without retraining, but existing methods pursue aggressive sparsity, where further
speedup costs disproportionately more attention fidelity. We target the opposite end
of this trade-off: fix near-lossless fidelity by construction, and remove as much
computation as this constraint permits. Two observations make this regime practical: roughly $40\%$ of
block interactions can be removed while retaining $99\%$ of the attention mass, and
the high-mass support remains stable across denoising steps. We propose \method{}, a training-free
sparse-attention method that fixes a retained-mass threshold of $99\%$ rather than a
sparsity ratio: it measures exact block attention masses at one early dense step,
keeps, for each head and query block, the smallest key/value block set meeting the
threshold, and
reuses the frozen block indices for all remaining steps. On Wan2.1-1.3B, \method{}
alone gives a $1.36\times$ speedup with a $0.06$-point VBench Overall drop. The
benefit is largest under composition: combined with feature caching, \method{}
reaches a $3.2\times$ speedup on HunyuanVideo at a $0.02$-point drop, versus $0.32$
points for the strongest sparse baseline at comparable speed. Across three video
diffusion transformers and speedups up to $3.2\times$, \method{} consistently
achieves the best training-free speed--quality trade-off.
\end{abstract}

\section{Introduction}
\label{sec:introduction}

Video diffusion transformers have become a leading architecture for text-to-video
generation~\citep{wan2025,kong2024hunyuanvideo,yang2024cogvideox}, but their sampling
cost remains high. Each denoising step applies self-attention over a long
3D token sequence, and because attention is quadratic in sequence
length, its share of the sampling cost grows with resolution and duration---from
roughly half of the latency at short contexts to more than $80\%$ at long
ones~\citep{xi2025svg}. Training-free acceleration is therefore especially
attractive, as it promises to lower inference cost without retraining the model,
changing the sampler, or sacrificing generation quality.

\begin{figure}[t]
\centering
\includegraphics[width=\linewidth]{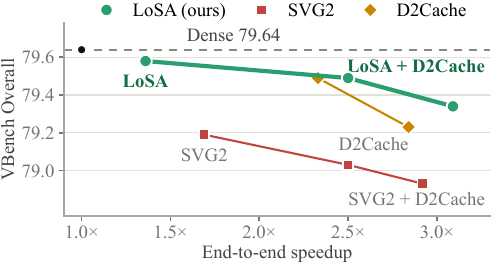}
\caption{Speed--quality trade-off on Wan2.1-1.3B. Each point reports VBench Overall
versus end-to-end speedup. \method{} and its cached variants form the entire Pareto
frontier: every baseline configuration is matched or dominated by a \method{}
configuration, showing that near-lossless sparse attention preserves quality better
than aggressive sparsity or cache-only acceleration.}
\label{fig:intro-tradeoff}
\end{figure}

Existing training-free accelerators reduce this cost along several complementary directions.
Feature caching reduces how often transformer blocks are computed, by reusing or
predicting intermediate outputs across denoising
steps~\citep{liu2025teacache,zhao2025pab,liu2026d2cache}. Sparse attention reduces the
cost of each computed attention layer by skipping less important query--key
interactions~\citep{xi2025svg,yang2025svg2,zhang2025spargeattn}. Distillation and
quantization are also effective~\citep{li2024t2vturbo,yin2024dmd2,zhao2025viditq,zhang2025sageattention},
but they require additional training, calibration, or model modification. In this work,
we focus on sparse attention as a drop-in inference-time replacement for dense
self-attention.

Most sparse-attention methods aim for high sparsity: they push the attention density as
low as possible while keeping visual quality acceptable---SVG2, for instance,
typically retains only $25$--$30\%$ of blocks~\citep{yang2025svg2}. This is a natural objective when sparse
attention is evaluated as a standalone speedup module. However, it also places the
method in an aggressive approximation regime, where additional sparsity can cost a
disproportionate amount of attention fidelity. The error becomes more consequential in a full
acceleration pipeline, where sparse attention is combined with feature caching:
cached or predicted states carry attention errors forward instead of confining them
to a single computation. Thus, low attention error is not merely a conservative
preference, but an important design criterion for composable inference acceleration.
Simply operating an existing method at a lower sparsity level does not necessarily
meet this criterion. Some methods fix a sparsity ratio, leaving the retained attention
mass unspecified; others rely on coarse mass estimates that are not sufficiently
accurate in the near-lossless regime. This leads to a different question: \textit{can
near-lossless attention be achieved by construction while reducing computational
cost enough to yield a meaningful speedup?}

Video diffusion attention satisfies both requirements. Its block-level distribution
contains a long low-mass tail, allowing substantial computation to be removed with
little mass loss, and after the rapidly changing early denoising steps, the high-mass
support remains stable enough to be constructed once and reused throughout the
remaining trajectory. These observations suggest that the right objective is not
maximum sparsity, but controlled, near-lossless sparsity.

We propose \method{}, a training-free, near-lossless replacement for self-attention
in video diffusion transformers. Instead of fixing a sparsity ratio, \method{} fixes a retained-mass
target. For each layer, head, and query block, it keeps the smallest set of key/value
blocks whose cumulative attention mass reaches a threshold $\theta$; we use
$\theta=0.99$ by default. \method{} builds this pattern once after a few dense denoising
steps and then freezes only the block indices. Because construction coincides with a
dense step, the selection is driven by exact block masses rather than the coarse
importance estimates that per-step sparse methods must rely on, so the retained-mass
target is met accurately rather than approximately. At every subsequent step, it
still recomputes Q/K/V and attention weights from the current hidden states, normalizing
the softmax over the retained keys. Thus, \method{} changes only the attention support,
not the model weights, denoising schedule, or cached features.

Empirically, \method{} consistently lies on the speed--quality Pareto frontier, both as
a standalone module and when combined with feature caching, as shown in
Figure~\ref{fig:intro-tradeoff}. Its advantage over more aggressive sparse-attention
configurations becomes more pronounced in cached pipelines, supporting our premise that
near-lossless sparse attention is a better primitive for composable inference
acceleration. This trend holds across the evaluated models and speedup levels.

Our contributions are as follows:
\begin{itemize}
    \item We identify and characterize a near-lossless sparse regime in video diffusion
    attention, where approximately $40\%$ of block interactions can be removed while
    retaining $99\%$ of dense attention mass.
    \item We introduce \method{}, which constructs a retained-mass support once per
    sample from exact block attention masses, independently for each layer, head, and
    query block, and freezes only the block indices for the remaining denoising steps.
    \item We demonstrate that \method{} is near-lossless as a standalone module and
    remains effective when combined with feature caching, improving the training-free
    speed--quality frontier across multiple models and speedup levels.
\end{itemize}

\begin{figure*}[t]
\centering
\includegraphics[width=\linewidth]{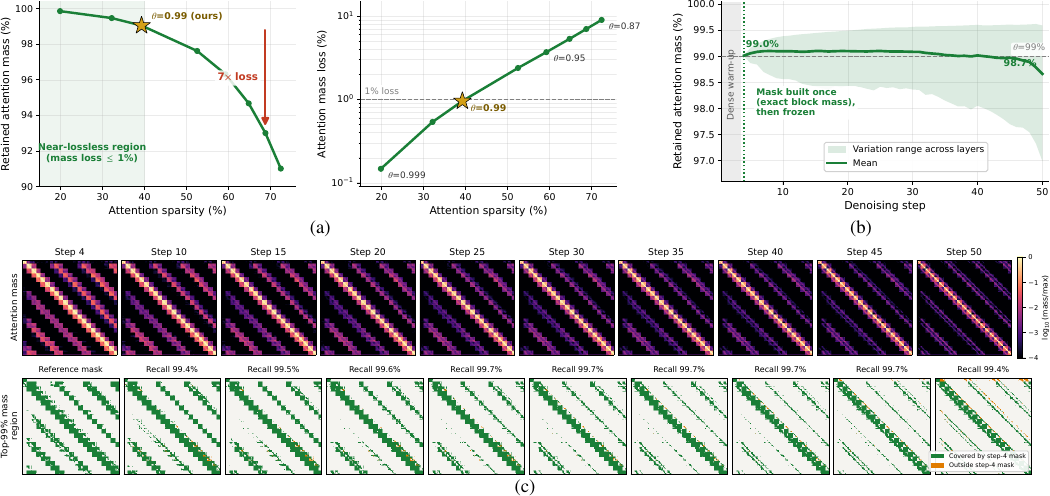}
\caption{Empirical motivation for near-lossless sparse attention. (a) Retained
attention mass versus attention sparsity (left) and the corresponding mass loss under
different thresholds $\theta$ (right). About $40\%$ of the block area can be removed
with only $1\%$ mass loss, while further sparsity becomes disproportionately expensive.
(b) Retained mass of a frozen $99\%$-mass pattern across denoising steps. The pattern is
constructed once after the dense warm-up and keeps its recall close to $99\%$ for the
rest of the trajectory. (c) Attention maps of one head across denoising steps (top, log scale) and each
step's top-99\%-mass region (bottom), colored by whether it is covered by the mask
built at $t_0=3$; the number above each panel is the covered fraction. The
high-mass region stays inside the early mask throughout denoising.}
\label{fig:motivation}
\end{figure*}

\section{Related Work}
\label{sec:related-work}

\subsection{Efficient Video Diffusion Inference}

Diffusion transformers have become a common backbone for text-to-video generation,
including Wan~\citep{wan2025}, HunyuanVideo~\citep{kong2024hunyuanvideo}, and
CogVideoX~\citep{yang2024cogvideox}. Compared with image generation, video generation
requires attention over much longer 3D token sequences, making inference
substantially more expensive. Existing acceleration methods can be roughly grouped by
where they reduce cost. Step distillation reduces the number of denoising
steps~\citep{li2024t2vturbo,yin2024dmd2}. Quantization reduces numerical precision and
memory traffic~\citep{zhao2025viditq,zhang2025sageattention,feng2026quantsparse}.
Caching reuses intermediate features across denoising steps, while sparse attention
reduces the cost of attention inside a computed step. Our work belongs to the last
category and is training-free.

\subsection{Sparse Attention for Video Generation}

Sparse attention exploits the fact that attention mass in video diffusion is highly
concentrated. Some methods use predefined layouts. Sliding Tile Attention uses a local
3D tile pattern~\citep{zhang2025sta}, Radial Attention adopts a radial
layout~\citep{li2025radial}, and DiTFastAttn studies static sparse patterns for
diffusion transformers~\citep{yuan2024ditfastattn}. These methods are simple and cheap
to apply, but a fixed layout may not match the content-dependent attention pattern of a
given prompt or denoising step.

Other methods build sparse patterns from the current attention or hidden states. Sparse
VideoGen selects spatial or temporal sparse masks for different attention
heads~\citep{xi2025svg}. Sparse VideoGen2 permutes tokens with semantic-aware
clustering so that important interactions become more
block-friendly~\citep{yang2025svg2}. SpargeAttn and XAttention estimate block
importance with lightweight online predictors~\citep{zhang2025spargeattn,xu2025xattention}.
Jenga dynamically removes less important tokens during generation~\citep{zhang2025jenga},
VSA learns sparse attention operators through training~\citep{zhang2025vsa}, and
AdaSpa also exploits cross-step pattern stability by reusing each searched mask
for several denoising steps~\citep{xia2025adaspa}. It uses this stability to
reduce the frequency of search, but still updates the mask at preset steps and
controls sparsity with a fixed budget. \method{} instead quantifies how well a
single pattern preserves attention mass and shows that one pattern selected by
exact retained mass can be reused for the remaining trajectory.

\subsection{Feature Caching}

Feature caching accelerates diffusion inference by skipping part of the computation
across denoising steps. DeepCache first shows that U-Net diffusion models contain
substantial reusable features~\citep{ma2024deepcache}. For diffusion transformers,
Pyramid Attention Broadcast shares attention outputs across nearby
steps~\citep{zhao2025pab}. Delta-DiT and FORA reuse feature deltas or layer outputs at
scheduled intervals~\citep{chen2024deltadit,selvaraju2024fora}. ToCa and DuCa cache
features at token granularity~\citep{zou2025toca,zou2024duca}. TeaCache, FasterCache,
and AdaCache decide when to reuse features using cheap difference or redundancy
signals~\citep{liu2025teacache,lv2024fastercache,kahatapitiya2025adacache}. D2Cache
uses second-order delta caching to further improve the speed--quality trade-off for
video diffusion~\citep{liu2026d2cache}.

Caching and sparse attention are complementary: the cache decides whether a step or
block is computed at all, while sparse attention lowers the cost of the attention
that is still computed, so the two can be combined. However, a cache may carry
attention errors from computed steps into later reused states, so the combination
favors sparse attention with low error. This motivates the near-lossless design of
\method{}.

\section{Motivation}
\label{sec:motivation}

This section presents the empirical basis of \method{}. We show that video diffusion
attention has a low-mass tail that can be removed with little mass loss, and
that its high-mass block pattern is stable enough to be constructed once and reused
across denoising steps. Unless otherwise stated, all measurements in this section are
averaged over ten prompts on Wan2.1-14B, across all self-attention layers and heads.

\subsection{Measuring Sparse Attention by Retained Mass}

We quantify the fidelity of a block-sparse support by the fraction of dense attention
mass it retains. For attention head $h$ at denoising step $t$ (the layer index is
omitted for brevity), let $A^h_t=\operatorname{softmax}(Q^h_t {K^h_t}^{\top}/\sqrt{d})$
be the dense attention probability matrix, where $d$ is the head dimension and the
softmax is applied row-wise. We partition query tokens into blocks $B^Q_i$ of size $R$ and key/value tokens into blocks $B^K_j$ of size $C$. The block-level attention contribution from query block $i$ to key/value block $j$ is
\begin{equation}
M^h_t(i,j)
=
\sum_{q\in B^Q_i}
\sum_{k\in B^K_j}
A^h_t(q,k).
\end{equation}
We refer to $M^h_t(i,j)$ as the block attention mass.

A block-sparse attention pattern keeps a subset $\Omega^h_i$ of key/value blocks for each head $h$ and query block $i$. We measure the dense attention mass retained by this pattern as
\begin{equation}
\operatorname{Recall}_t(\Omega)
=
\frac{
\sum_h \sum_i \sum_{j\in\Omega^h_i} M^h_t(i,j)
}{
\sum_h \sum_i \sum_j M^h_t(i,j)
}.
\end{equation}
We measure the remaining block computation by the coverage
$\operatorname{Coverage}(\Omega)=(1/H)\sum_h \sum_i |\Omega^h_i|/(n_q n_k)$,
where $H$ is the number of heads and $n_q,n_k$ are the numbers of query and key/value
blocks. A near-lossless sparse pattern should have recall close to one while reducing
coverage meaningfully.

\subsection{Near-Lossless Sparsity Is the Cost-Effective Regime}

Figure~\ref{fig:motivation}(a) evaluates the relationship between retained mass and removed block area. For each query block, key/value blocks are ordered by their block attention mass, and we measure how much total mass remains as more low-mass blocks are removed.

The curve shows a strong diminishing-return effect. Removing the low-mass tail incurs
little mass loss: dropping roughly 40\% of the block area reduces the retained mass by
only about 1\%. However, pushing much further is considerably more expensive. At roughly
70\% sparsity, the mass loss grows to about 7\%, seven times larger. The right panel of
Figure~\ref{fig:motivation}(a) shows the same effect from the perspective of the
retained-mass threshold: tightening it to $\theta=0.999$ removes only about 20\% of
the block area, while relaxing it to $\theta=0.95$ saves more computation but loses
several times more mass than $\theta=0.99$.

These results motivate optimizing retained mass directly rather than prescribing a
fixed sparsity ratio: the trade-off for near-lossless acceleration lies earlier on
the curve than where aggressive methods operate. \method{} therefore fixes a
retained-mass threshold, choosing $\theta=0.99$ inside the near-lossless region
identified in Figure~\ref{fig:motivation}(a).

\subsection{Near-Lossless Patterns Are Stable Across Denoising}

The near-lossless regime reduces attention cost, but its savings per step are
moderate by design, so pattern construction must be cheap and accurate. A pattern
built once at an early dense step satisfies both requirements, as long as it remains
valid at later steps.

Figure~\ref{fig:motivation}(b) tests one-time construction directly. At step $t_0=3$,
after three dense denoising steps, we construct a block pattern that retains 99\% of
the block attention mass at that step. We then freeze this pattern and
evaluate its recall using the true dense attention mass at subsequent steps. The mean
retained mass stays close to 99\% throughout the remaining trajectory and is still
98.7\% at the final step; even the worst-performing layers retain more than 97\% of the
mass. A near-lossless pattern therefore remains near-lossless across the subsequent
denoising trajectory.

Figure~\ref{fig:motivation}(c) shows why. Attention starts broad and then
concentrates: the top-99\% region of every later step is almost a subset of the
region at $t_0$. The support contracts within an early envelope rather than
migrating, so a mask built after the initial transition keeps covering the dominant
regions without being updated.

\begin{figure}[t]
\centering
\includegraphics[width=\linewidth]{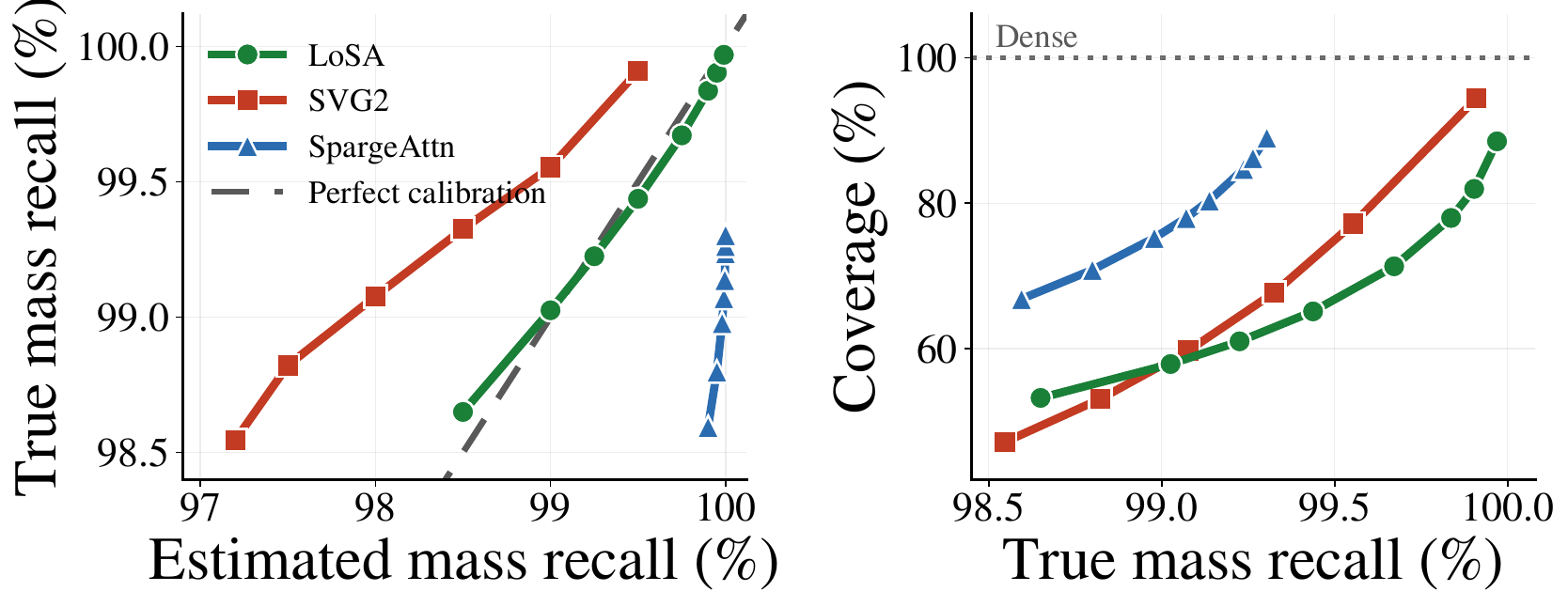}
\caption{\method{} achieves more accurate and efficient attention-mass
estimation in the near-lossless regime. The left panel plots estimated against true
retained-mass recall as each method's fidelity setting is varied. The pattern
constructed by \method{} at $t_0$ remains close to perfect calibration, whereas
the per-step estimates used by SVG2 and SpargeAttn are miscalibrated. The right
panel shows the coverage required to achieve each true recall.
In the near-lossless regime at or above $99\%$ recall, \method{} consistently
requires less coverage than both SVG2 and SpargeAttn.}
\label{fig:estimation}
\end{figure}

A natural alternative is to raise the thresholds of existing top-$p$
estimation methods to target the same near-lossless regime. However, methods
such as SVG2 and SpargeAttn rely on coarse mass estimates that are not
sufficiently accurate in this regime. Figure~\ref{fig:estimation} confirms
this: \method{} remains close to the perfect-calibration line in the
high-recall region (left) and requires less coverage than both baselines at
true recall of $99\%$ or higher (right), even before accounting for their
per-step estimation overhead.

Together, these observations lead to the core design of \method{}. After the
initial attention transition, \method{} constructs a sparse pattern from exact
block masses at $t_0$ and reuses it for the rest of denoising. This preserves
near-lossless recall while avoiding repeated estimation.

\section{Methodology}
\label{sec:methodology}

\begin{figure*}[t]
\centering
\includegraphics[width=\linewidth]{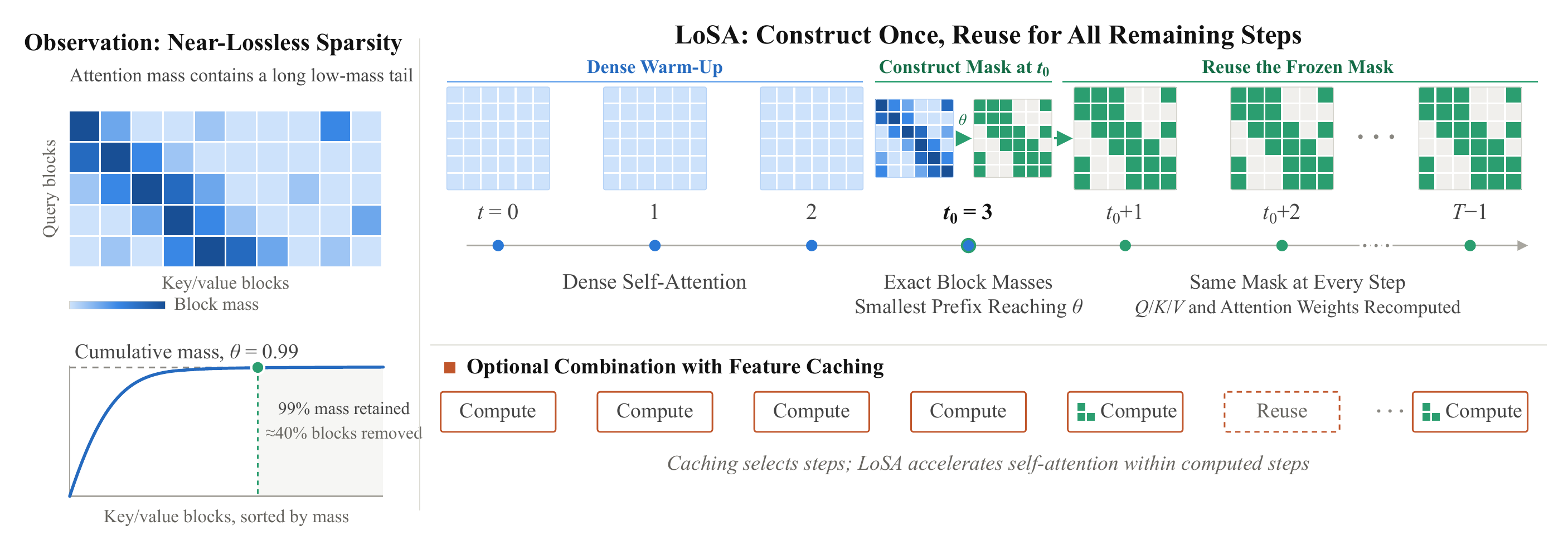}
\caption{Overview of \method{}. Left: video diffusion attention contains a low-mass
tail, allowing approximately $40\%$ of the block area to be removed while retaining
approximately $99\%$ of the attention mass. Right: after a dense warm-up, \method{}
measures exact block masses at $t_0$ and constructs the mask $\Omega$ by retaining,
for each layer, head, and query block, the smallest prefix whose cumulative mass
reaches $\theta$. The same frozen mask is reused at all remaining steps, while Q/K/V
and attention weights are recomputed from the current hidden states. Bottom: with
optional feature caching, the cache selects which steps are computed, while \method{}
accelerates self-attention within every computed step.}
\label{fig:overall}
\end{figure*}

\method{} replaces self-attention in video diffusion transformers without modifying
model weights, prompts, sampling schedules, or classifier-free guidance settings. Figure~\ref{fig:overall} connects the empirical
premise of \method{} to its inference procedure. \method{} operates in two stages: it
constructs a mass-preserving sparse support once online for each sample and reuses the
frozen support during the remaining denoising steps. No offline calibration or profiling
is required. Unless otherwise stated, we use $\theta=0.99$, run dense attention at steps
$0,1,2$, construct the support during the dense step $t_0=3$, and use
frozen-support sparse attention afterward.

\subsection{One-Time Pattern Construction}

At the construction step $t_0$, \method{} computes the block attention masses
$M^h_{t_0}(i,j)$ defined in the Motivation section for every self-attention layer,
head, query block, and key/value block. For each head $h$ and query block $i$, let
$\pi^h_i$ sort the key/value blocks by decreasing mass, with ties broken arbitrarily.
Since each query row of $A^h_{t_0}$ sums to one, the total mass of a query block of
size $R$ is $R$, and the retained-mass threshold $\theta$ translates into the shortest
prefix
\begin{equation}
s^h_i
=
\min \Big\{
m \in \{1,\ldots,n_k\}:
\sum_{r=1}^{m} M^h_{t_0}(i,\pi^h_i(r)) \ge \theta R
\Big\}.
\end{equation}
\method{} keeps $\Omega^h_i = \{\pi^h_i(1), \ldots, \pi^h_i(s^h_i)\}$, stored
independently for each layer, head, and query block, and separately for the
conditional and unconditional branches under classifier-free guidance. Since
$\Omega$ only records block indices, its memory footprint is orders of magnitude
smaller than that of a token-level mask.

Because construction coincides with a dense attention step, the masses
$M^h_{t_0}(i,j)$ are exact rather than estimated. This gives the threshold a
direct meaning: setting $\theta=0.99$ retains $99\%$ of the mass at $t_0$ by
construction. This guarantee is exact at $t_0$; at later steps, near-lossless
recall is maintained by the cross-step support stability shown in
Figure~\ref{fig:motivation}(b--c). Figure~\ref{fig:estimation} compares this exact construction with
approximate top-$p$ estimation. \method{} stays close to perfect calibration in
the high-recall region, whereas SVG2 and SpargeAttn show larger gaps between
estimated and true recall.

\subsection{Sparse Attention with a Frozen Pattern}

\method{} uses dense self-attention up to and including the construction step $t_0$.
Afterward, each query attends only to its retained keys: for a token $q$ in query
block $i(q)$, let $\mathcal{K}^h(q)=\bigcup_{j\in\Omega^h_{i(q)}} B^K_j$. The output
of head $h$ at step $t$ is
\begin{equation}
O^h_t(q)
=
\sum_{k\in\mathcal{K}^h(q)}
\frac{\exp\!\big(Q^h_t(q)\cdot K^h_t(k)/\sqrt{d}\big)}
{\sum_{k'\in\mathcal{K}^h(q)}\exp\!\big(Q^h_t(q)\cdot K^h_t(k')/\sqrt{d}\big)}\,
V^h_t(k),
\end{equation}
with Q/K/V computed from the current hidden states.

Since discarded keys are removed from the softmax denominator, \method{} is not
mathematically identical to dense attention; it is near-lossless in practice because
$\Omega$ preserves nearly all block attention mass. We do not impose auxiliary local
or diagonal blocks: the retained-mass criterion directly preserves whichever blocks
dominate the dense distribution.

For one head, dense self-attention costs $O(n_q n_k RCd)$ and frozen sparse attention costs $O(\sum_i |\Omega^h_i| RCd)$, where $d$ is the head dimension. Thus, after the one-time construction step, the attention cost is reduced in proportion to the coverage defined in the Motivation section.

\subsection{Combination with Feature Caching}

\method{} and feature caching act on different parts of inference. The cache decides whether a denoising step or transformer block is computed, reused, or predicted; whenever a block is computed, \method{} replaces its dense self-attention with frozen-support sparse attention. We do not change the cache threshold, reuse schedule, predictor, or cached residuals. In our combination with D2Cache, support construction is scheduled on a fully computed warm-up step, ensuring that the dense block masses required to build $\Omega$ are available. After construction, skipped steps or blocks simply reuse cached states, while computed blocks use the same frozen support as in standalone inference.

\begin{table*}[t]
    \centering
    \setlength{\tabcolsep}{3pt}
    {\small
    \begin{tabular}{@{}lcc*{8}{c}@{}}
    \toprule
    \multicolumn{1}{c}{Method} & Time (s) $\downarrow$ & Speedup $\uparrow$ &
    \multicolumn{5}{c}{VBench Dimensions} &
    \multicolumn{3}{c}{Aggregate Scores} \\
    \cmidrule(lr){4-8}\cmidrule(l){9-11}
    & & & Flicker $\uparrow$ & Dynamic $\uparrow$ & Aesthetic $\uparrow$ &
    Scene $\uparrow$ & Consistency $\uparrow$ & Quality $\uparrow$ &
    Overall $\uparrow$ & $\Delta\uparrow$ \\
    \midrule
    
    \multicolumn{11}{@{}l}{\textbf{Wan2.1-T2V-1.3B, 480p}} \\
    \rowcolor[gray]{0.95}
    Dense & 104 & $1.00\times$ & 99.72 & 72.22 & 57.34 & 31.10 & 25.03 & 82.45 & 79.64 & -- \\
    
    \midrule
    \multicolumn{11}{@{}l}{\hspace{0.7em}\textit{Sparse attention only}} \\
    SpargeAttn & 80 & $1.30\times$ & 99.10 & 70.83 & 53.58 & 28.34 & 24.78 & 79.51 & 76.80 & $-2.84$ \\
    SVG1 & 60 & $\mathbf{1.75\times}$ & 99.25 & \textbf{75.00} & 54.07 & 25.00 & 24.69 & 80.35 & 77.63 & $-2.01$ \\
    SVG2 & 62 & $1.69\times$ & 99.65 & 70.83 & 56.71 & 31.32 & 24.78 & 81.75 & 79.19 & $-0.45$ \\
    \textbf{\method{}} & 77 & $1.36\times$ & \textbf{99.74} & 70.83 & \textbf{57.31} & \textbf{35.17} & \textbf{24.93} & \textbf{82.19} & \textbf{79.58} & $\mathbf{-0.06}$ \\
    
    \hdashline
    \multicolumn{11}{@{}l}{\hspace{0.7em}\textit{$\approx2.5\times$ Speedup}} \\
    D2Cache & 45 & $2.33\times$ & 99.71 & 69.44 & 57.14 & 31.61 & 24.81 & 82.20 & \textbf{79.49} & $\mathbf{-0.15}$ \\
    SVG2 + D2Cache & 42 & $\mathbf{2.50\times}$ & 99.64 & 70.83 & 56.84 & 30.60 & 24.76 & 81.75 & 79.03 & $-0.61$ \\
    \textbf{\method{} + D2Cache} & 42 & $\mathbf{2.50\times}$ & \textbf{99.73} & \textbf{72.22} & \textbf{57.32} & \textbf{32.99} & \textbf{24.83} & \textbf{82.24} & \textbf{79.49} & $\mathbf{-0.15}$ \\
    
    \hdashline
    \multicolumn{11}{@{}l}{\hspace{0.7em}\textit{$\approx3\times$ Speedup}} \\
    D2Cache & 37 & $2.84\times$ & 99.71 & 70.83 & 56.73 & 29.07 & 24.71 & 82.08 & 79.23 & $-0.41$ \\
    SVG2 + D2Cache & 36 & $2.92\times$ & 99.64 & \textbf{72.22} & 56.64 & 29.72 & 24.71 & 81.83 & 78.93 & $-0.71$ \\
    \textbf{\method{} + D2Cache} & 34 & $\mathbf{3.09\times}$ & \textbf{99.73} & \textbf{72.22} & \textbf{56.97} & \textbf{31.18} & \textbf{24.90} & \textbf{82.10} & \textbf{79.34} & $\mathbf{-0.30}$ \\
    
    \specialrule{\heavyrulewidth}{4pt}{2pt}
    \multicolumn{11}{@{}l}{\textbf{Wan2.1-T2V-14B, 720p}} \\
    \rowcolor[gray]{0.95}
    Dense & 2000 & $1.00\times$ & 98.97 & 86.11 & 60.39 & 33.94 & 26.20 & 82.51 & 81.02 & -- \\
    
    \midrule
    \multicolumn{11}{@{}l}{\hspace{0.7em}\textit{$\approx2.5\times$ Speedup}} \\
    SVG2 + D2Cache & 812 & $2.46\times$ & 98.97 & 86.11 & 59.50 & 33.50 & 26.01 & 82.00 & 80.31 & $-0.71$ \\
    \textbf{\method{} + D2Cache} & 801 & $\mathbf{2.50\times}$ & \textbf{99.06} & \textbf{87.50} & \textbf{59.89} & \textbf{35.17} & \textbf{26.08} & \textbf{82.36} & \textbf{80.59} & $\mathbf{-0.43}$ \\
    
    \specialrule{\heavyrulewidth}{4pt}{2pt}
    \multicolumn{11}{@{}l}{\textbf{HunyuanVideo-13B, 540p}} \\
    \rowcolor[gray]{0.95}
    Dense & 672 & $1.00\times$ & 98.61 & 79.17 & 60.72 & 29.51 & 26.37 & 83.29 & 79.66 & -- \\
    
    \midrule
    \multicolumn{11}{@{}l}{\hspace{0.7em}\textit{$\approx3.2\times$ Speedup}} \\
    SVG2 + D2Cache & 211 & $3.18\times$ & 98.49 & \textbf{79.17} & \textbf{60.74} & 28.27 & 26.44 & 82.85 & 79.34 & $-0.32$ \\
    \textbf{\method{} + D2Cache} & 211 & $\mathbf{3.19\times}$ & \textbf{98.54} & 77.78 & 60.60 & \textbf{29.14} & \textbf{26.47} & \textbf{82.94} & \textbf{79.64} & $\mathbf{-0.02}$ \\
    
    \bottomrule
    \end{tabular}%
    }
    \caption{End-to-end efficiency and generation quality on Wan2.1 and HunyuanVideo, with
    methods grouped by inference setting at approximately matched speed. $\Delta$ is the
    change in VBench Overall relative to the corresponding dense baseline (gray rows);
    $\Delta\uparrow$ indicates that a smaller drop is better.
    Latencies are averaged over the full VBench prompt suite.
    Bold marks the highest speedup or the best quality within each group; ties after
    rounding are both highlighted. All scores are percentages; ``Consistency'' denotes
    VBench overall consistency.}
    \label{tab:main-results}
    \end{table*}
    
    \section{Experiments}
    \label{sec:experiments}
    
    \subsection{Experimental Setup}
    
    \paragraph{Models and benchmark.}
    We evaluate \method{} on three text-to-video diffusion transformers:
    Wan2.1-T2V-1.3B at 480p, Wan2.1-T2V-14B at 720p~\citep{wan2025}, and
    HunyuanVideo-13B at 540p~\citep{kong2024hunyuanvideo}, all with their
    official checkpoints and default sampling configurations.
    Generation quality is measured with VBench~\citep{huang2024vbench} on its
    full standard prompt suite; given the cost of large-scale video generation,
    each configuration is evaluated with a single fixed random seed. We report
    five representative VBench dimensions (temporal flickering, dynamic degree,
    aesthetic quality, scene, and overall consistency), together with
    the aggregate Quality and Overall scores. Each method is summarized by its
    Overall drop $\Delta$ relative to the corresponding dense baseline.
    
    \paragraph{Baselines and operating points.}
    Our sparse-attention baselines are SVG1~\citep{xi2025svg},
    SVG2~\citep{yang2025svg2}, and SpargeAttn~\citep{zhang2025spargeattn}, all
    training-free. We do not include AdaSpa~\citep{xia2025adaspa} because its
    code is unavailable and its reported HunyuanVideo performance is below
    SVG2, our strongest reproducible baseline. Its reported operating point is
    also comparable to our fixed-sparsity ablation
    (Table~\ref{tab:ablation-rule}). On the caching side we use
    D2Cache~\citep{liu2026d2cache}, a state-of-the-art method, and we also
    evaluate the sparse--cache combinations. Since quality is only comparable
    at equal speed, we compare methods at operating points with matched
    end-to-end speedup: about $2.5\times$ and $3\times$ on Wan2.1-1.3B,
    $2.5\times$ on Wan2.1-14B, and $3.2\times$ on HunyuanVideo. Each operating
    point is reached by adjusting only the reuse schedule of D2Cache. All
    sparse-attention methods keep their default configurations throughout, and
    SVG1 and SpargeAttn appear only in the standalone comparison.
    
    \paragraph{Implementation.}
    All experiments run on NVIDIA H200 GPUs in a Diffusers-based inference
    environment.
    \method{} is applied to self-attention only, since cross-attention accounts
    for a much smaller fraction of inference cost in our target models. All
    layers use query block size $R=128$ and key/value block size $C=32$, the
    finest granularity that did not compromise block-sparse kernel efficiency
    in our tests. We use $\theta=0.99$ and $t_0=3$ as defaults. Block-sparse
    attention is implemented with FlashInfer, and the construction step uses a
    custom dense-attention kernel that accumulates exact block masses while
    computing full attention. Construction adds a one-time overhead of roughly
    one denoising step, which is included in all reported latencies and
    amortized over the remaining sparse steps.
    
    \subsection{Main Results}
    
    Table~\ref{tab:main-results} reports end-to-end latency and VBench quality
    for all models and operating points, and Figure~\ref{fig:intro-tradeoff}
    visualizes the resulting speed--quality frontier on Wan2.1-1.3B.
    Qualitative comparisons of the generated videos are provided in the
    supplementary material.
    
    \paragraph{Standalone sparse attention.}
    On Wan2.1-1.3B, \method{} alone accelerates sampling by $1.36\times$ while
    reducing VBench Overall by only $0.06$ points, staying close to dense
    attention on every reported dimension. SVG2 reaches a higher standalone
    speedup ($1.69\times$) but loses $0.45$ points, over seven times the
    degradation. The earlier SVG1 ($1.75\times$) and SpargeAttn ($1.30\times$)
    lose $2.01$ and $2.84$ points, and SpargeAttn is even slower than
    \method{}; we therefore treat SVG2 as the aggressive reference in the
    remaining comparisons. \method{} deliberately trades part of the
    standalone speedup for near-losslessness.
    
    \paragraph{Combination with feature caching.}
    The advantage of near-losslessness shows most clearly under composition.
    At the $\approx2.5\times$ operating point, D2Cache alone reaches
    $2.33\times$ with a $0.15$-point drop. Composing it with SVG2 raises the
    speedup to $2.50\times$ but increases the quality loss to $0.61$ points,
    worse than the cache alone: the attention error of the sparse module is
    written into cached states and propagated across reused steps, so the
    combination loses more quality than either component. Composing the same
    cache with \method{} reaches the same $2.50\times$ with the quality drop
    unchanged at $0.15$ points. In effect, \method{} multiplies the speedup of
    the cache at no measurable quality cost. At $\approx3\times$, the
    contrast sharpens: \method{}+D2Cache is simultaneously the fastest
    ($3.09\times$) and the highest-quality ($\Delta=-0.30$) configuration,
    surpassing both D2Cache alone ($2.84\times$, $-0.41$) and SVG2+D2Cache
    ($2.92\times$, $-0.71$).
    
    \paragraph{Larger model and different backbone.}
    The advantage persists at scale. On the two larger models we evaluate only
    the composed, high-speedup setting: a single dense sample takes hundreds to
    thousands of seconds, so a full grid of operating points is prohibitively
    expensive, and high speedups are also the practically relevant setting at
    this scale. On Wan2.1-14B at 720p, where a dense sample
    takes roughly $2000$ seconds, \method{}+D2Cache achieves $2.50\times$ with a
    $0.43$-point drop, against $2.46\times$ and $0.71$ points for SVG2+D2Cache.
    On HunyuanVideo-13B, the margin widens further: at a matched
    $\approx3.2\times$ speedup, \method{}+D2Cache is virtually lossless
    ($\Delta=-0.02$) while SVG2+D2Cache loses $0.32$ points. The near-lossless
    regime is thus not specific to a single model family.
    
    \subsection{Ablation Studies}
    
    The ablations examine the two design choices of \method{}: the selection
    rule and the retained-mass threshold. Both act directly on retained
    attention mass, so we evaluate them by recall rather than benchmark
    scores. This is also cheap: each variant is just a different selection
    rule applied to the same recorded masses, so no additional generation is
    needed. As a reference point, the default configuration's $99.0\%$ average
    recall corresponds to its $0.06$-point Overall drop in
    Table~\ref{tab:main-results}. Following the measurement protocol of the
    Motivation section, masses are recorded on Wan2.1-1.3B over $100$ prompts
    across all layers, heads, and denoising steps; latencies are measured
    end-to-end on the same prompts.
    
    \begin{table}[t]
    \centering
    \setlength{\tabcolsep}{3.5pt}
    {\small
    \begin{tabular}{@{}lccc@{}}
    \toprule
    Selection rule & Coverage (\%) & \multicolumn{2}{c}{Recall (\%) $\uparrow$} \\
    \cmidrule(l){3-4}
     & & Average & Worst layer \\
    \midrule
    Fixed-ratio top-$k$ & 66.1 & 96.5 & 86.6 \\
    Retained-mass ($\theta=0.99$) & 66.1 & \textbf{99.0} & \textbf{97.3} \\
    \bottomrule
    \end{tabular}
    }
    \caption{Comparison of selection rules at equal average coverage on
    Wan2.1-1.3B ($100$ prompts). With coverage matched, both rules run equally
    fast ($1.37\times$ end-to-end), but fixed-ratio top-$k$ loses much more
    attention mass, especially in its worst layer. Recall is averaged over the denoising trajectory;
    ``Worst layer'' is the layer with the lowest average recall.}
    \label{tab:ablation-rule}
    \end{table}

    \begin{table}[t]
    \centering
    \setlength{\tabcolsep}{4.5pt}
    {\small
    \begin{tabular}{@{}lcccc@{}}
    \toprule
     & Dense & $\theta=0.999$ & $\theta=0.99$ & $\theta=0.95$ \\
    \midrule
    Coverage (\%) & 100 & 83.5 & 66.1 & 46.6 \\
    Recall (\%) & 100 & 99.9 & 99.0 & 95.9 \\
    Time (s) & 104.8 & 83.1 & 76.3 & 69.0 \\
    Saved (s) & -- & 21.7 & 28.5 & 35.8 \\
    Speedup & $1.00\times$ & $1.26\times$ & $1.37\times$ & $1.52\times$ \\
    \bottomrule
    \end{tabular}
    }
    \caption{Effect of the retained-mass threshold $\theta$ on Wan2.1-1.3B
    ($100$ prompts). The realized recall stays close to the target at every
    setting, and the savings diminish quickly: most of the time reduction is
    already obtained at the default $\theta=0.99$.}
    \label{tab:ablation-theta}
    \end{table}
    
    \paragraph{Retained mass vs.\ a fixed sparsity ratio.}
    Some block-sparse methods use a fixed sparsity level shared across
    heads~\citep{zhang2025spargeattn,xia2025adaspa}. However, attention
    concentration varies widely from head to head: some heads concentrate their
    mass on a few blocks, while others spread it broadly. \method{} therefore
    selects blocks by cumulative mass instead of a fixed ratio. To verify this
    choice, we compare against a fixed-ratio top-$k$ variant matched to the same average coverage
    (Table~\ref{tab:ablation-rule}). It runs equally fast but retains only
    $96.5\%$ of the mass, more than three times the default's loss, and its
    worst-layer recall falls to $86.6\%$ against $97.3\%$: the uniform budget
    truncates exactly those heads whose attention is spread broadly.
    
    \paragraph{Choice of $\theta$.}
    The remaining design choice is the threshold $\theta$, \method{}'s only
    tuning parameter. Two things need checking: whether setting $\theta$
    actually delivers the requested recall, and why $0.99$ is the right
    default. Table~\ref{tab:ablation-theta} confirms the first: the realized
    recall never falls more than $0.1$ points below the target (and sits above
    it at $\theta=0.95$), so $\theta$ is a reliable fidelity control. The
    default then follows from the sharply nonlinear trade between recall and
    time: giving up the first point of mass ($\theta=0.99$) saves $28.5$
    seconds, while giving up three more points ($\theta=0.95$) recovers only
    $7.3$ more. We therefore default to $\theta=0.99$, the knee of this
    trade-off.

\section{Conclusion}
\label{sec:conclusion}

We identified a near-lossless sparse regime in video diffusion attention:
roughly $40\%$ of block interactions can be removed while retaining $99\%$ of
the attention mass, and the high-mass support is stable enough to be
constructed once and reused. \method{} exploits this regime with a
deliberately simple design: one dense step yields exact block masses, the
threshold $\theta$ specifies the retained mass directly, and every subsequent
step reuses the frozen block indices. As a standalone module, \method{}
delivers a meaningful speedup while remaining nearly lossless. Composed with
feature caching, it multiplies the speedup of a strong cache at almost no
additional quality cost, whereas the aggressive sparse-attention baseline
degrades the combination below the cache alone. The result is the best training-free
speed--quality trade-off across all evaluated models and operating points.
We hope near-lossless sparse attention becomes a standard primitive for
composable diffusion inference acceleration.

\bibliography{aaai2027}

@article{wan2025,
  title  = {{Wan}: Open and Advanced Large-Scale Video Generative Models},
  author = {{Team Wan} and Wang, Ang and Ai, Baole and Wen, Bin and Mao, Chaojie and Xie, Chen-Wei and Chen, Di and Yu, Feiwu and Zhao, Haiming and Yang, Jianxiao and Zeng, Jianyuan and Wang, Jiayu and Zhang, Jingfeng and Zhou, Jingren and Wang, Jinkai and Chen, Jixuan and Zhu, Kai and Zhao, Kang and Yan, Keyu and Huang, Lianghua and Feng, Mengyang and Zhang, Ningyi and Li, Pandeng and Wu, Pingyu and Chu, Ruihang and Feng, Ruili and Zhang, Shiwei and Sun, Siyang and Fang, Tao and Wang, Tianxing and Gui, Tianyi and Weng, Tingyu and Shen, Tong and Lin, Wei and Wang, Wei and Wang, Wei and Zhou, Wenmeng and Wang, Wente and Shen, Wenting and Yu, Wenyuan and Shi, Xianzhong and Huang, Xiaoming and Xu, Xin and Kou, Yan and Lv, Yangyu and Li, Yifei and Liu, Yijing and Wang, Yiming and Zhang, Yingya and Huang, Yitong and Li, Yong and Wu, You and Liu, Yu and Pan, Yulin and Zheng, Yun and Hong, Yuntao and Shi, Yupeng and Feng, Yutong and Jiang, Zeyinzi and Han, Zhen and Wu, Zhi-Fan and Liu, Ziyu},
  journal = {arXiv preprint arXiv:2503.20314},
  year   = {2025}
}

@article{kong2024hunyuanvideo,
  title  = {{HunyuanVideo}: A Systematic Framework For Large Video Generative Models},
  author = {Kong, Weijie and Tian, Qi and Zhang, Zijian and Min, Rox and Dai, Zuozhuo and Zhou, Jin and Xiong, Jiangfeng and Li, Xin and Wu, Bo and Zhang, Jianwei and Wu, Kathrina and Lin, Qin and Yuan, Junkun and Long, Yanxin and Wang, Aladdin and Wang, Andong and Li, Changlin and Huang, Duojun and Yang, Fang and Tan, Hao and Wang, Hongmei and Song, Jacob and Bai, Jiawang and Wu, Jianbing and Xue, Jinbao and Wang, Joey and Wang, Kai and Liu, Mengyang and Li, Pengyu and Li, Shuai and Wang, Weiyan and Yu, Wenqing and Deng, Xinchi and Li, Yang and Chen, Yi and Cui, Yutao and Peng, Yuanbo and Yu, Zhentao and He, Zhiyu and Xu, Zhiyong and Zhou, Zixiang and Xu, Zunnan and Tao, Yangyu and Lu, Qinglin and Liu, Songtao and Zhou, Dax and Wang, Hongfa and Yang, Yong and Wang, Di and Liu, Yuhong and Jiang, Jie and Zhong, Caesar},
  journal = {arXiv preprint arXiv:2412.03603},
  year   = {2024}
}

@inproceedings{yang2024cogvideox,
  title     = {{CogVideoX}: Text-to-Video Diffusion Models with An Expert Transformer},
  author    = {Yang, Zhuoyi and Teng, Jiayan and Zheng, Wendi and Ding, Ming and Huang, Shiyu and Xu, Jiazheng and Yang, Yuanming and Hong, Wenyi and Zhang, Xiaohan and Feng, Guanyu and Yin, Da and Zhang, Yuxuan and Wang, Weihan and Cheng, Yean and Xu, Bin and Gu, Xiaotao and Dong, Yuxiao and Tang, Jie},
  booktitle = {International Conference on Learning Representations (ICLR)},
  year      = {2025}
}

@inproceedings{xi2025svg,
  title     = {Sparse Video-Gen: Accelerating Video Diffusion Transformers with Spatial-Temporal Sparsity},
  author    = {Xi, Haocheng and Yang, Shuo and Zhao, Yilong and Xu, Chenfeng and Li, Muyang and Li, Xiuyu and Lin, Yujun and Cai, Han and Zhang, Jintao and Li, Dacheng and Chen, Jianfei and Stoica, Ion and Keutzer, Kurt and Han, Song},
  booktitle = {Proceedings of the 42nd International Conference on Machine Learning},
  series    = {Proceedings of Machine Learning Research},
  volume    = {267},
  pages     = {68208--68224},
  publisher = {PMLR},
  year      = {2025},
  url       = {https://proceedings.mlr.press/v267/xi25c.html}
}

@inproceedings{yang2025svg2,
  title     = {Sparse VideoGen2: Accelerate Video Generation with Sparse Attention via Semantic-Aware Permutation},
  author    = {Yang, Shuo and Xi, Haocheng and Zhao, Yilong and Li, Muyang and Zhang, Jintao and Cai, Han and Lin, Yujun and Li, Xiuyu and Xu, Chenfeng and Peng, Kelly and Chen, Jianfei and Han, Song and Keutzer, Kurt and Stoica, Ion},
  booktitle = {Advances in Neural Information Processing Systems (NeurIPS)},
  year      = {2025},
  note      = {arXiv:2505.18875}
}

@inproceedings{zhang2025spargeattn,
  title     = {{SpargeAttention}: Accurate and Training-free Sparse Attention Accelerating Any Model Inference},
  author    = {Zhang, Jintao and Xiang, Chendong and Huang, Haofeng and Wei, Jia and Xi, Haocheng and Zhu, Jun and Chen, Jianfei},
  booktitle = {Proceedings of the 42nd International Conference on Machine Learning},
  series    = {Proceedings of Machine Learning Research},
  volume    = {267},
  pages     = {76397--76413},
  publisher = {PMLR},
  year      = {2025},
  url       = {https://proceedings.mlr.press/v267/zhang25ch.html}
}

@inproceedings{zhang2025sta,
  title     = {Fast Video Generation with Sliding Tile Attention},
  author    = {Zhang, Peiyuan and Chen, Yongqi and Su, Runlong and Ding, Hangliang and Stoica, Ion and Liu, Zhengzhong and Zhang, Hao},
  booktitle = {Proceedings of the 42nd International Conference on Machine Learning},
  series    = {Proceedings of Machine Learning Research},
  volume    = {267},
  pages     = {74714--74731},
  publisher = {PMLR},
  year      = {2025},
  url       = {https://proceedings.mlr.press/v267/zhang25m.html}
}

@inproceedings{yuan2024ditfastattn,
  title     = {DiTFastAttn: Attention Compression for Diffusion Transformer Models},
  author    = {Yuan, Zhihang and Zhang, Hanling and Lu, Pu and Ning, Xuefei and Zhang, Linfeng and Zhao, Tianchen and Yan, Shengen and Dai, Guohao and Wang, Yu},
  booktitle = {Advances in Neural Information Processing Systems (NeurIPS)},
  year      = {2024},
  note      = {arXiv:2406.08552}
}

@inproceedings{liu2025teacache,
  title     = {Timestep Embedding Tells: It's Time to Cache for Video Diffusion Model},
  author    = {Liu, Feng and Zhang, Shiwei and Wang, Xiaofeng and Wei, Yujie and Qiu, Haonan and Zhao, Yuzhong and Zhang, Yingya and Ye, Qixiang and Wan, Fang},
  booktitle = {Proceedings of the IEEE/CVF Conference on Computer Vision and Pattern Recognition (CVPR)},
  pages     = {7353--7363},
  month     = jun,
  year      = {2025}
}

@inproceedings{zhao2025pab,
  title     = {Real-Time Video Generation with Pyramid Attention Broadcast},
  author    = {Zhao, Xuanlei and Jin, Xiaolong and Wang, Kai and You, Yang},
  booktitle = {International Conference on Learning Representations (ICLR)},
  year      = {2025},
  note      = {arXiv:2408.12588}
}

@inproceedings{liu2026d2cache,
  title     = {{D2Cache}: Second-Order Delta Caching for Higher Video Diffusion Acceleration},
  author    = {Liu, Enhuai and Wang, Yunke and Sun, Changming and Xu, Chang},
  booktitle = {Proceedings of the IEEE/CVF Conference on Computer Vision and Pattern Recognition (CVPR)},
  pages     = {43589--43599},
  month     = jun,
  year      = {2026}
}

@inproceedings{huang2024vbench,
  title     = {{VBench}: Comprehensive Benchmark Suite for Video Generative Models},
  author    = {Huang, Ziqi and He, Yinan and Yu, Jiashuo and Zhang, Fan and Si, Chenyang and Jiang, Yuming and Zhang, Yuanhan and Wu, Tianxing and Jin, Qingyang and Chanpaisit, Nattapol and Wang, Yaohui and Chen, Xinyuan and Wang, Limin and Lin, Dahua and Qiao, Yu and Liu, Ziwei},
  booktitle = {Proceedings of the IEEE/CVF Conference on Computer Vision and Pattern Recognition (CVPR)},
  pages     = {21807--21818},
  month     = jun,
  year      = {2024}
}

@inproceedings{li2024t2vturbo,
  title     = {T2V-Turbo: Breaking the Quality Bottleneck of Video Consistency Model with Mixed Reward Feedback},
  author    = {Li, Jiachen and Feng, Weixi and Fu, Tsu-Jui and Wang, Xinyi and Basu, Sugato and Chen, Wenhu and Wang, William Yang},
  booktitle = {Advances in Neural Information Processing Systems (NeurIPS)},
  year      = {2024},
  note      = {arXiv:2405.18750}
}

@inproceedings{yin2024dmd2,
  title     = {Improved Distribution Matching Distillation for Fast Image Synthesis},
  author    = {Yin, Tianwei and Gharbi, Micha{\"e}l and Park, Taesung and Zhang, Richard and Shechtman, Eli and Durand, Fr{\'e}do and Freeman, William T.},
  booktitle = {Advances in Neural Information Processing Systems (NeurIPS)},
  year      = {2024},
  note      = {arXiv:2405.14867}
}

@inproceedings{zhao2025viditq,
  title     = {ViDiT-Q: Efficient and Accurate Quantization of Diffusion Transformers for Image and Video Generation},
  author    = {Zhao, Tianchen and Fang, Tongcheng and Huang, Haofeng and Wan, Rui and Soedarmadji, Widyadewi and Liu, Enshu and Li, Shiyao and Lin, Zinan and Dai, Guohao and Yan, Shengen and Yang, Huazhong and Ning, Xuefei and Wang, Yu},
  booktitle = {International Conference on Learning Representations (ICLR)},
  year      = {2025}
}

@inproceedings{zhang2025sageattention,
  title     = {SageAttention: Accurate 8-Bit Attention for Plug-and-play Inference Acceleration},
  author    = {Zhang, Jintao and Wei, Jia and Zhang, Pengle and Zhu, Jun and Chen, Jianfei},
  booktitle = {International Conference on Learning Representations (ICLR)},
  year      = {2025}
}

@inproceedings{xia2025adaspa,
  title     = {Training-free and Adaptive Sparse Attention for Efficient Long Video Generation},
  author    = {Xia, Yifei and Ling, Suhan and Fu, Fangcheng and Wang, Yujie and Li, Huixia and Xiao, Xuefeng and Cui, Bin},
  booktitle = {Proceedings of the IEEE/CVF International Conference on Computer Vision (ICCV)},
  pages     = {15982--15993},
  month     = oct,
  year      = {2025}
}

@inproceedings{li2025radial,
  title     = {Radial Attention: $O(n\log n)$ Sparse Attention with Energy Decay for Long Video Generation},
  author    = {Li, Xingyang and Li, Muyang and Cai, Tianle and Xi, Haocheng and Yang, Shuo and Lin, Yujun and Zhang, Lvmin and Yang, Songlin and Hu, Jinbo and Peng, Kelly and Agrawala, Maneesh and Stoica, Ion and Keutzer, Kurt and Han, Song},
  booktitle = {Advances in Neural Information Processing Systems (NeurIPS)},
  year      = {2025}
}

@inproceedings{xu2025xattention,
  title     = {{XAttention}: Block Sparse Attention with Antidiagonal Scoring},
  author    = {Xu, Ruyi and Xiao, Guangxuan and Huang, Haofeng and Guo, Junxian and Han, Song},
  booktitle = {Proceedings of the 42nd International Conference on Machine Learning},
  series    = {Proceedings of Machine Learning Research},
  volume    = {267},
  pages     = {69819--69831},
  publisher = {PMLR},
  year      = {2025},
  url       = {https://proceedings.mlr.press/v267/xu25ag.html}
}

@inproceedings{zhang2025vsa,
  title     = {VSA: Faster Video Diffusion with Trainable Sparse Attention},
  author    = {Zhang, Peiyuan and Chen, Yongqi and Huang, Haofeng and Lin, Will and Liu, Zhengzhong and Stoica, Ion and Xing, Eric and Zhang, Hao},
  booktitle = {Advances in Neural Information Processing Systems (NeurIPS)},
  year      = {2025},
  note      = {arXiv:2505.13389}
}

@inproceedings{zhang2025jenga,
  title     = {Training-Free Efficient Video Generation via Dynamic Token Carving},
  author    = {Zhang, Yuechen and Xing, Jinbo and Xia, Bin and Liu, Shaoteng and Peng, Bohao and Tao, Xin and Wan, Pengfei and Lo, Eric and Jia, Jiaya},
  booktitle = {Advances in Neural Information Processing Systems (NeurIPS)},
  year      = {2025},
  note      = {arXiv:2505.16864}
}

@article{chen2024deltadit,
  title   = {{$\Delta$-DiT}: A Training-Free Acceleration Method Tailored for Diffusion Transformers},
  author  = {Chen, Pengtao and Shen, Mingzhu and Ye, Peng and Cao, Jianjian and Tu, Chongjun and Bouganis, Christos-Savvas and Zhao, Yiren and Chen, Tao},
  journal = {arXiv preprint arXiv:2406.01125},
  year    = {2024}
}

@article{selvaraju2024fora,
  title   = {FORA: Fast-Forward Caching in Diffusion Transformer Acceleration},
  author  = {Selvaraju, Pratheba and Ding, Tianyu and Chen, Tianyi and Zharkov, Ilya and Liang, Luming},
  journal = {arXiv preprint arXiv:2407.01425},
  year    = {2024}
}

@inproceedings{kahatapitiya2025adacache,
  title     = {Adaptive Caching for Faster Video Generation with Diffusion Transformers},
  author    = {Kahatapitiya, Kumara and Liu, Haozhe and He, Sen and Liu, Ding and Jia, Menglin and Zhang, Chenyang and Ryoo, Michael S. and Xie, Tian},
  booktitle = {Proceedings of the IEEE/CVF International Conference on Computer Vision (ICCV)},
  pages     = {15240--15252},
  month     = oct,
  year      = {2025}
}

@inproceedings{zou2025toca,
  title     = {Accelerating Diffusion Transformers with Token-wise Feature Caching},
  author    = {Zou, Chang and Liu, Xuyang and Liu, Ting and Huang, Siteng and Zhang, Linfeng},
  booktitle = {International Conference on Learning Representations (ICLR)},
  year      = {2025},
  note      = {arXiv:2410.05317}
}

@inproceedings{ma2024deepcache,
  title     = {DeepCache: Accelerating Diffusion Models for Free},
  author    = {Ma, Xinyin and Fang, Gongfan and Wang, Xinchao},
  booktitle = {Proceedings of the IEEE/CVF Conference on Computer Vision and Pattern Recognition (CVPR)},
  pages     = {15762--15772},
  month     = jun,
  year      = {2024}
}

@inproceedings{lv2024fastercache,
  title     = {FasterCache: Training-Free Video Diffusion Model Acceleration with High Quality},
  author    = {Lv, Zhengyao and Si, Chenyang and Song, Junhao and Yang, Zhenyu and Qiao, Yu and Liu, Ziwei and Wong, Kwan-Yee K.},
  booktitle = {International Conference on Learning Representations (ICLR)},
  year      = {2025},
  note      = {arXiv:2410.19355}
}

@article{zou2024duca,
  title   = {Rethinking Token-wise Feature Caching: Accelerating Diffusion Transformers with Dual Feature Caching},
  author  = {Zou, Chang and Zheng, Shikang and Zhang, Evelyn and Guo, Runlin and Xu, Haohang and Shi, Zhengyi and He, Conghui and Hu, Xuming and Zhang, Linfeng},
  journal = {IEEE Transactions on Image Processing},
  volume  = {35},
  pages   = {6211--6220},
  year    = {2026},
  doi     = {10.1109/TIP.2026.3698363}
}

@inproceedings{feng2026quantsparse,
  title     = {QuantSparse: Comprehensively Compressing Video Diffusion Transformer with Model Quantization and Attention Sparsification},
  author    = {Feng, Weilun and Yang, Chuanguang and Qin, Haotong and Wu, Mingqiang and Li, Yuqi and Li, Xiangqi and An, Zhulin and Huang, Libo and Zhang, Yulun and Magno, Michele and Xu, Yongjun},
  booktitle = {International Conference on Learning Representations (ICLR)},
  year      = {2026},
  note      = {arXiv:2509.23681}
}

\end{document}